# Physics-informed Deep Learning to Solve Three-dimensional Terzaghi's Consolidation Equation: Forward and Inverse Problems

Mr Biao Yuan[a], PhD Candidate

cnby@leeds.ac.uk

Dr Ana Heitor[a], Lecturer

A.Heitor@leeds.ac.uk

Dr He Wang[b], Associate Professor

he_wang@ucl.ac.uk

Dr Xiaohui Chen[a*], Associate Professor

Corresponding author

X.Chen@leeds.ac.uk

a. Geomodelling and Artificial Intelligence Centre, School of Civil Engineering, University of Leeds, Leeds, LS2 9JT, UK

b. Department of Computer Science, Faculty of Engineering Sciences, University College London, London, WC1E 6BT, UK

**Abstract:** The emergence of neural networks constrained by physical governing equations has sparked a new trend in deep learning research, which is known as Physics-Informed Neural Networks (PINNs). However, solving high-dimensional problems with PINNs is still a substantial challenge, the space complexity brings difficulty to solving large multidirectional problems. In this paper, a novel PINN framework to quickly predict several three-dimensional Terzaghi's consolidation cases under different conditions is proposed. Meanwhile, the loss functions for different cases are introduced, and their differences in three-dimensional consolidation problems are highlighted. The tuning strategies for the PINNs framework for three-dimensional consolidation problems are introduced. Then, the performance of PINNs is tested and compared with traditional numerical methods adopted in forward problems, and the coefficients of consolidation and the impact of noisy data in inverse problems are identified. Finally, the results are summarized and presented from three-dimensional simulations of PINNs, which show an accuracy rate of over 99% compared with ground truth for both forward and inverse problems. These results are desirable with good accuracy and can be used for soil settlement prediction, which demonstrates that the proposed PINNs framework can learn the three-dimensional consolidation PDE well.

**Keywords:** Three-dimensional Terzaghi's consolidation; Physics-informed neural networks (PINNs); Forward problems; Inverse problems; soil settlement

# 1. Introduction

Soil consolidation and settlement are critical problems in geotechnical engineering and occur during the construction of large infrastructures such as highways, railways, and airports (Lo et al., 2016, Pham et al., 2019). The theory of soil consolidation describes the excess pore water pressure dissipation in porous media. The pioneering works of Terzaghi et al. (1996) and Biot (2004) have made significant contributions, starting from the initial formulation for a 1D case and later extending it to encompass three-dimensional cases. Schiffman and Stein (1970) studied the consolidation equation under a linear increase in load with time and derived an analytical solution for one-dimensional soil consolidation in this scenario. In addition, Analytical solutions are available for the governing equation in one-dimensional cases (Verruijt, 2013). The high-dimensional Terzaghi's consolidation equations can be much more challenging to solve analytically than the one-dimensional cases, as they involve more complex variables and boundary conditions. But solving high-dimensional Terzaghi's consolidation equations is essential for predicting the behavior of soil layers under a wide range of loading conditions (Lo et al., 2016, Pham et al., 2019), such as statistic loading, earthquake loading, excavation, and embankment construction. For complex or high-dimensional forward problems, numerical methods such as finite difference methods and finite element methods are generally applied (Lewis and Schrefler, 1987). Traditional numerical methods have matured for several decades, but they can be time-consuming and computationally complicated when solving high-dimensional equations.

Physics-informed deep learning is an increasingly emerging field that combines the strengths of both physics systems and deep learning to solve partial differential equations (PDEs), which utilizes neural networks as a flexible and powerful tool to approximate the solution of PDEs. Unlike traditional numerical methods that rely on discretizing the problem domain and approximating the solution at discrete points (Gao et al., 2021), physics-informed deep learning is a mesh-free method, which operates directly on the continuous domain and trains the neural network to fit the physical constraints over the entire domain (Karniadakis et al., 2021, Lu et al., 2021, Raissi et al., 2019). One advantage of physics-informed deep learning is that it leverages the underlying physical system to underpin the neural network algorithms. In this way, a physically meaningful solution by incorporating the governing equations as constraints in the training process can be achieved (Raissi et al., 2019). Another advantage of physics-informed deep learning is that it can handle various PDEs and conditions with a small training dataset (Zeng et al., 2022). Moreover, physics-informed deep learning has an absolute advantage in inverse problems such as parameter identification and equation inversion, which may be not easy to directly solve by traditional methods (Raissi et al., 2019,

Long et al., 2019, Mao et al., 2020, Raissi et al., 2018, Raissi and Karniadakis, 2018, Chen et al., 2020, Chen et al., 2021). The parallelization capability of deep learning models (Shukla et al., 2021, Meng et al., 2020) allows for the efficient computation of solutions on high-performance computing platforms. Therefore, it is possible to tackle complex and computationally intensive problems that would otherwise be intractable by using traditional numerical methods (Karniadakis et al., 2021). In recent years, physics-informed deep learning theory has been reported in past studies. It has played a significant role in modeling and simulating physical systems in various fields such as solid mechanics, fluid mechanics, heat transfer, and quantum mechanics, et al. (Cai et al., 2021b, Raissi et al., 2019, Bai et al., 2023, Cai et al., 2021a, Cai et al., 2021c, Diao et al., 2023, Guo et al., 2023, Haghighat et al., 2021, Hua et al., 2023, Koric and Abueidda, 2023, Mao et al., 2020, Raissi et al., 2018). For instance, Karimpouli and Tahmasebi (2020) compared the Gaussian process (GP) and physics-informed neural networks for solving a 1-dimensional (1D) time-dependent seismic wave equation. Both methods showed high accuracy with a smaller amount of training data, highlighting the effectiveness of these meshless approaches. Mao et al. (2020) approximated the Euler equations for high-speed aerodynamic flows and highlighted the superiority of PINNs over traditional numerical methods in solving inverse problems that would otherwise be challenging with standard techniques, although some limitations in terms of accuracy were highlighted for forward problems. Cai et al. (2021b) introduced two different DeepM&Mnet architectures employing deep operator networks (DeepONets) as blocks, and these architectures demonstrate accurate and efficient inference of 2D electroconvection fields based on unseen electric potentials. Zhang et al. (2021) provided a method of identifying the best meteorological input variables for a deep Recurrent Neural Network (RNN) model to predict daily runoff. Two different deep RNN models, a long-short-term memory (LSTM) model, and a gated recurrent unit (GRU) model, were comparatively applied to predict runoff with these inputs.

The successful application of PINNs in many fields makes it possible to explore their potential application for soil consolidation. PINNs can incorporate the underlying physics of the problem, such as Darcy's law and the principle of conservation of mass, into the model learning process, which can efficiently lead to more accurate solutions with the physics constraints (Zhang, 2022). In addition, PINNs serve as promising surrogate approaches for solving PDEs efficiently and accurately and usually have better performance and convenience in inverse problems (Karniadakis et al., 2021). For example, Bekele (2021) and Zhang et al. (2022) adopted PINNs to solve one-dimensional soil consolidation cases in both forward and inverse problems. Zhang et al. (2023) developed a physics-informed data-driven approach to identify governing equations from experimental data and then solve them in one-dimensional soil consolidation. Lu and Mei (2022) proposed a physics-based deep learning method to predict two-dimensional soil consolidation. However, few studies are aimed at three-dimensional soil consolidation problems with PINNs. Due to the curse of dimensionality and big training and testing data, it is hard for PINNs to get stable convergence and fine accuracy in high-dimensional problems. To this end, PINN application analysis specifically for three-dimensional consolidation problems needs to be solved urgently.

This paper focuses on the study of physics-informed deep learning for three-dimensional soil consolidation with four independent variables, namely three spatial coordinates and a temporal coordinate. The governing equation, physics-informed neural network, and reference solutions are first introduced briefly. Then, the PINN method is adopted to solve the three-dimensional consolidation problems and make predictions under different conditions, and comparative analyses are made between the PINN solutions and the reference solutions subsequently. Furthermore, the coefficients of consolidation and soil settlement are predicted by PINNs in three-dimensional cases, and the sensitivity analysis and noisy data influence are also presented. Finally, a brief conclusion is given, and the current challenges and future perspectives of PINNs are discussed.

## 2. Three-dimensional Terzaghi's Consolidation and Governing Equation
### 2.1 Terzaghi's Consolidation Theory

Terzaghi (1925) introduced his consolidation theory for soil in his seminal work on soil mechanics, which formed an integral part of his comprehensive theory of soil mechanics. Biot (2004) developed his consolidation theory considering the relationship between pore pressure and soil skeleton deformation during soil consolidation by incorporating the effective stress principle, soil continuity, and equilibrium equation. Terzaghi's Consolidation Theory is a critical concept in geotechnical engineering, which describes the consolidation process of the soil layer under static load (Terzaghi, 1925, Terzaghi et al., 1996). The soil layer is typically assumed as homogeneous and isotropic, and the consolidation process is governed by Darcy's law and the principle of conservation of mass. Terzaghi's consolidation theory relies on the use of nonlinear second-order partial differential equations to describe the change in excess pore water pressure and displacement of the soil layer over time during consolidation. The equations take the form of the diffusion type, where the changing rate of pore water pressure is proportional to the rate of change of displacement. The constant of proportionality is known as the coefficient of consolidation ($C_v$) and is dependent on the properties of the soil, such as its permeability and compressibility. The solutions of these equations underpin modern FEM simulations, and the analytical approach still provides reasonable estimations of long-term settlement, which provide insights into the consolidation process, including the time required for the soil layer to reach a state of equilibrium and the magnitude of settlement that can be expected.

### 2.2 Governing Equations

The governing equation for three-dimensional Terzaghi's consolidation theory (Biot, 2004)(Figure 1) is given below:

$$\frac{\partial u}{\partial t} = C_v\left(\frac{\partial^2 u}{\partial x^2} + \frac{\partial^2 u}{\partial y^2} + \frac{\partial^2 u}{\partial z^2}\right) \tag{a}$$

where $u$ is the excess pore water pressure, $C_v$ is the coefficient of consolidation, $t$ represents time, and $x, y, z$ are the spatial coordinates. The initial and boundary conditions can be typically established as follows:

$$\begin{cases} u|_{(x,y,z)\in I} = p_0(t=0) \\ u|_{(x,y,z)\in J} = 0(t>0) \\ \frac{\partial u}{\partial n}\bigg|_{(x,y,z)\in J} = 0(t>0) \end{cases} \tag{b}$$

where $p_0$ is the initial pressure of the soil layer, $I$ is the spatial coordinate domain except for boundaries, $J$ is the boundary domain, and $n$ is the specific coordinate among $x, y$ and $z$. Equations (a) and (b) provide the initial condition, Dirichlet and Neumann boundary conditions for three-dimensional consolidation problems under different cases.

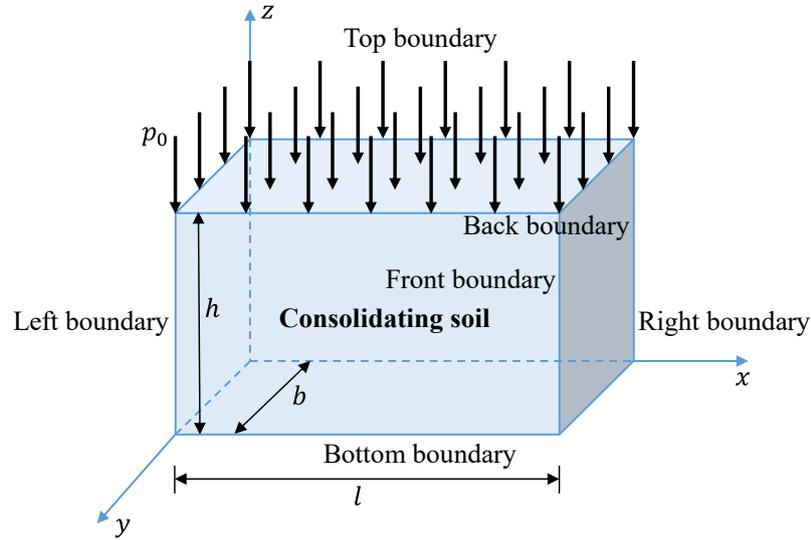

Figure 1 Three-dimensional Terzaghi's consolidation.

As shown in Figure 1, the three-dimensional Terzaghi's consolidation is a high-dimensional and multidirectional problem. The three-dimensional cubic consolidation model is selected for this study, $l$ and $b$ represent the length of the consolidation model while $h$ indicates the height of the cubic consolidating soil. The three-dimensional cubic consolidation model has 6 boundaries, namely the top and bottom boundaries, left and right boundaries, and front and back boundaries, making the spatial boundary conditions and force conditions change in multiple dimensions and directions. In general, traditional numerical methods are computationally complex and inefficient when studying three-dimensional and multidirectional soil consolidation. In this study, a uniformly distributed load $p_0$ is considered to act on the top boundary of the three-dimensional cubic consolidating soil, and multidirectional boundary conditions are considered to verify the effectiveness and efficiency of the proposed PINNs method for three-dimensional consolidation problems.

### 3. Physics-informed Neural Network

In this section, the neural network architecture for three-dimensional problems is first discussed, along with the automatic differentiation which is one of the most important features applied in the PINNs for solving three-dimensional partial differential equations. Next, the loss function for three-dimensional consolidation problems is defined and presented. Then, the hyper-parameters, which play an important role in model training, are also specifically reviewed. Finally, the Python environment developed in this study is presented.

### 3.1 Neural Network Architecture

The deep neural network used in this study is the fully connected neural network, which is a type of artificial neural network (ANN). In a fully connected neural network, the hidden layers are the layers except the input and output layers. While the hidden units in a single hidden layer are neurons, each neuron in a layer is connected to every neuron in the former and next layers (Schwing and Urtasun, 2015). This means each neuron in a layer receives input from all the neurons in the former layer and sends output to all the neurons in the latter layer. Layers are connected with activation functions such as Sigmoid, ReLU, and Tanh (Sharma et al., 2017, Agostinelli et al., 2014, Karlik and Olgac, 2011, Nwankpa et al., 2018). The Tanh function is adopted in this study. The weights and biases of the neurons are adjusted during model training by using optimization algorithms such as stochastic gradient descent (SGD), Limited-memory Broyden–Fletcher–Goldfarb–Shanno (L-BFGS) and Adaptive Moment Estimation (Adam), allowing the model to learn the connection between the input and output data by minimizing the loss function (Le et al., 2011, Zaheer and Shaziya, 2019). The L-BFGS algorithm, adopted in this study, is a quasi-Newton and full-batch gradient-based optimization algorithm.

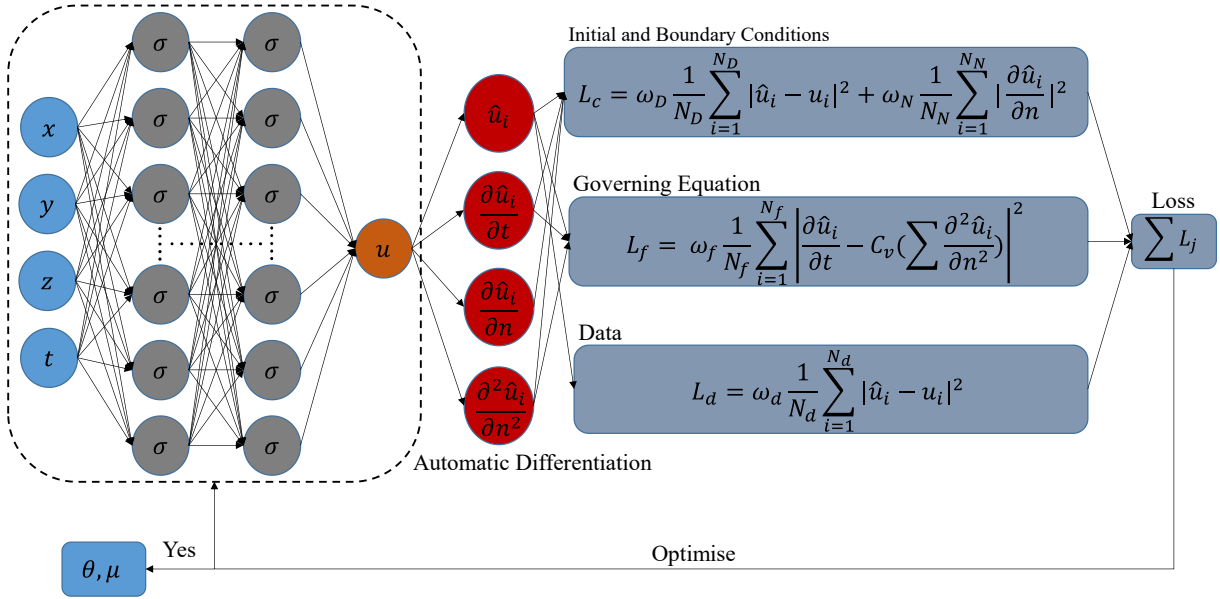

**Figure 2** The architecture of the Physics-informed Neural Network (PINN) with a Fully Connected Neural Network with input, hidden and output Layers for three-dimensional Terzaghi's consolidation equation with spatial and temporal coordinates $(x, y, z, t)$: Note: The activation function employed at the hidden layers is $\text{Tanh}(x)$. Automatic Differentiation (AD) is adopted to compute the partial derivatives in the governing equation that serves as a physical constraint for optimization together. The number of hidden layers and neurons in this figure is for illustrative purposes only, and the actual number of hidden layers and neurons utilized for different cases is discussed in relevant parts.

In a fully connected neural network (Figure 2), the input data is typically flattened into a one-dimensional vector and fed into the input layer. Each neuron in the first layer applies a linear transformation to the input vector and a non-linear activation function (hyperbolic tangent: Tanh) is used to produce an output. The output of each neuron in the former layer serves as input to each neuron in the next layer until the final output layer produces a solution or prediction. The weights and biases of a fully connected neural network are learned through process-denominated backpropagation. In this process, the gradient of the loss function which is related to the network parameters is computed and used to adjust and update the weights and biases adaptively. In this paper, the L-BFGS algorithm is used to optimize the loss function so that the error is minimized. However, the neural network may occasionally encounter the problem of overfitting if the network is too large or the training data is insufficient. In such cases, techniques such as dropout and regularization can be employed to mitigate this challenge.

The fully connected neural network employs the feedforward mechanism to approximate the solution of three-dimensional PDEs. In general, the PDE governing the real physics of the problem takes the form below:

$$u_t + N[u] = 0, x, y, z \in \Omega, t \in [0, T], \tag{c}$$

where $u(x, y, z, t)$ denotes the unknown hidden solution of the three dimensional PDE, $N[\cdot]$ is a nonlinear differential operator, $\Omega$ and $T$ are the whole spatial and temporal domain respectively. The value of each neural in the feedforward neural network can be represented as follows:

$$u_{i,j} = \sigma(W_{i-1,j} X_{i-1} + b_{i-1,j}) \tag{d}$$

where $u_{i,j}$ indicates the approximated value of the $j$-th neuron in the $i$-th layer; $\sigma(\cdot)$ is the activation function employed in the hidden layers; $W_{i-1,j}$ denotes the $j$-th group of weight vector of the $(i-1)$-th layer, while $b_{i-1,j}$ is the $j$-th bias of the $(i-1)$-th layer. $X_{i-1}$ represents the input vector of $i$-th layer composed of the output values of all neurons in the $(i-1)$-th layer. Therefore, the output result of the entire neural network with $n$ layers can be expressed as follows:

$$u_n = W_{n-1} \cdot \sigma(\cdots \sigma(W_3 \cdot \sigma(W_2 \cdot \sigma(W_1 X_1 + b_1) + b_2) + b_3) \cdots) + b_{n-1} \tag{e}$$

According to the uniform approximation principle of the neural network, the output result $u_n$ of the neural network is used to approximate the latent solution $u(x, y, z, t)$ of the three dimensional PDE to satisfy Equation (c):

$$u(x, y, z, t) \approx u_n \tag{f}$$

$$(u_n)_t + N[u_n] = 0 \tag{g}$$

In the following sections, how the neural network works to better satisfy Equation (g) will be discussed.

### 3.2 Loss Function

The Mean Squared Error (MSE) loss function is a commonly used metric as a physical constraint in PINNs to evaluate the difference and error between the predicted output and the exact values over training points and collocation points. The use of the MSE loss function provides a clear and intuitive measure of how well the model performs. The goal of training a PINN model is to ensure the MSE loss function converges towards zero (no error between ground truths and training outputs), which is achieved by adjusting the weights and biases of the model using an optimization algorithm (L-BFGS). For consolidation problems, the total loss function is defined as follows:

$$MSE = \omega_u MSE_u + \omega_f MSE_f \qquad (h)$$

Where $\omega_u$ and $\omega_f$ are the loss weights parameters (set to 1), $MSE_f$ is the loss of PDE residuals and $MSE_u$ represents the loss from the initial and Dirichlet boundary conditions and can be defined as follows:

$$MSE_u = \frac{1}{N_u} \sum_{i=1}^{N_u} |\hat{u}_i - u_i|^2 \qquad (i)$$

In Equation (i), $N_u$ is the number of training points in initial and boundary conditions, $\hat{u}_i$ is the approximated solution of the PDEs predicted by PINNs, and $u_i$ is the exact values of the PDEs.

The Neumann boundary conditions are also considered in this study. For this case $MSE_u$ can be represented as follows:

$$MSE_u = \frac{1}{N_u} \sum_{i=1}^{N_u} \left|\frac{\partial \hat{u}_i}{\partial n}\right|^2 \qquad (j)$$

The loss of PDEs residuals $MSE_f$ is given in Equation (k).

$$MSE_f = \frac{1}{N_f} \sum_{i=1}^{N_f} \left|\frac{\partial \hat{u}_i}{\partial t} - C_v \left(\sum \frac{\partial^2 \hat{u}_i}{\partial n^2}\right)\right|^2 \qquad (k)$$

The right side of Equation (k) is the deformation calculated using the PDE representing the consolidation process, which is expected to have a value of zero. $N_f$ is the number of collocation points in the defined domain, $C_v$ is the coefficient of consolidation, $n$ indicates the spatial coordinates $x, y, z$, and the values of partial derivatives are computed by automatic differentiation. In different dimensional cases, Equation (h) has variant forms. The loss function for three-dimensional problems contains more loss terms and a large number of data points, and thus conversion and balance of the loss function is more challenging. The details for different consolidation cases will be discussed in the corresponding sections.

### 3.3 Hyper-Parameters and Model Training

Hyper-parameters, which are set before training and are not learned from the data, are crucial for determining the accuracy and efficiency of PINNs. They often include variables such as the number of hidden layers, the number of neurons in each layer, learning rate, and batch size (Smith, 2018). For example, the number of hidden layers and the number of neurons in each layer can have significant impacts on the model's ability to capture the underlying physics of the problem. Moreover, setting the learning rate too high may make the model difficult to converge to the optimal point and result in poor performance, while setting it too low may cause the training process to become too slow. The suitable batch size can increase the speed and extent of model optimization. However, the evaluation of the optimal set of hyper-parameters for a PINN can be challenging, as it requires a significant amount of experimentation and tuning. In addition to the traditional hyper-parameters, there are also specialized hyper-parameters for PINNs that are related to the physics-informed constraints. These include the weights given to the physics-informed loss function relative to the data-driven loss function, and the parameters that control the regularization of the solution to ensure that it satisfies the governing equations of the problem. The setting of these hyperparameters will be presented in the corresponding sections.

PINNs are trained on a dataset that includes both input-output pairs and constraints based on the underlying physical principles of the problem. The input-output pairs in this study are obtained from real values and simulation data. For consolidation problems, the number of training points ($N_u$) is randomly selected in the initial and boundary conditions, while the number of collocation points ($N_f$) is randomly sampled from the whole domain by adopting Latin Hypercube Sampling (LHS), which is a typically uniform sampling method. The output predicted by the model represents $\hat{u}^i$ following the physical principle of the partial differential equations and the reference solution ($u_i$) will be introduced in section 4. The constraints, on the other hand, are used to guide the training of the neural network, ensuring that the resulting solution satisfies the underlying physics.

### 3.4 Automatic Differentiation and Python Environment

As shown in Figure 2, the partial derivatives should be computed before the loss function is optimized, that is where Automatic Differentiation (AD) comes into play (Griewank, 1989). In PINNs, gradients are used to update the weights and biases of the neural network as well as the loss function calculation during the model training, which is a crucial step in improving the neural network's performance. Automatic differentiation can be used to compute the gradients of the loss function related to the network parameters automatically, without the need for manual derivation or numerical approximation (Baydin et al., 2018). The technique is based on the chain rule of calculus, which allows the decomposition of the gradient of a composite function into the product of gradients of the individual functions in the chain (Paszke et al., 2017).

In this study, the model training was performed on an NVIDIA Tesla T4 GPU, and for the hyper-parameter combinations used in this study, the three-dimensional model training time for 5,000 epochs was about 30 min under this environment. As expected, models with lower numbers of hidden layers, hidden units, and epochs take a shorter time for training. For PINNs, the automatic differentiation capability in Pytorch is utilized (Paszke et al., 2017). PyTorch is a Python-based open-source machine learning library used for training and developing deep learning models (Paszke et al., 2019), which is developed by Facebook's AI research team and provides an easy-to-use platform for establishing neural networks and other machine learning models.

## 4. Reference Solution for Comparisons

In order to test the accuracy and effectiveness of the proposed PINN method, an error evaluation of the PINN model is required. Here, the numerical solution is used as a reference for three-dimensional consolidation problems. In the following sections, comprehensive comparisons between the PINN solution $\hat{u}_i$ and the reference solution $u_i$ will be implemented, purposefully, to quantify the difference between these two solutions.

### 4.1 Numerical Solution

Several numerical methods can be employed to solve three-dimensional consolidation equations, including Finite Element Methods (FEMs) (Johnson, 2012) and Finite Difference Methods (FDMs) (Morton and Mayers, 2005). These methods encompass a range of numerical techniques such as the Newton iteration method, Euler method, Modified Euler method, Runge-Kutta method, and Adams method (Morton and Mayers, 2005, Johnson, 2012, Lapidus and Pinder, 2011). FEM software packages such as ABAQUS, COMSOL, MATLAB, or Python can be employed for mathematical modelling. In this study, the second-order central difference method is employed to obtain the numerical solution for Terzaghi's consolidation equations (Lapidus and Pinder, 2011).

Equation (a) is a deformation problem with $x$, $y$, and $z$ as the coordinate axis, first separate the entire solidification area in the $x$, $y$, and $z$ directions into equal distances $\Delta x$, $\Delta y$ and $\Delta z$ grid, while the time is also divided into small segments $\Delta t$, then the left side of the equation can be approximated by rewriting the forward differential format, as follows:

$$\frac{\partial u}{\partial t} = \frac{u(x,y,z,t+\Delta t) - u(x,y,z,t)}{\Delta t} \tag{l}$$

The right-hand side of the equation can be approximated using the following differential form:

$$\frac{\partial^2 u}{\partial x^2} = \frac{u(x+\Delta x,y,z,t) - 2u(x,y,z,t) + u(x-\Delta x,y,z,t)}{\Delta x^2} \tag{m}$$

$$\frac{\partial^2 u}{\partial y^2} = \frac{u(x,y+\Delta y,z,t) - 2u(x,y,z,t) + u(x,y-\Delta y,z,t)}{\Delta y^2} \tag{n}$$

$$\frac{\partial^2 u}{\partial z^2} = \frac{u(x,y,z+\Delta z,t)-2u(x,y,z,t)+u(x,y,z-\Delta z,t)}{\Delta z^2} \tag{o}$$

Therefore, the differential form of Equation (a) can be expressed by Equation (p).

$$\frac{u(x,y,z,t+\Delta t)-u(x,y,z,t)}{\Delta t} = C_v \left( \frac{u(x+\Delta x,y,z,t)-2u(x,y,z,t)+u(x-\Delta x,y,z,t)}{\Delta x^2} + \frac{u(x,y+\Delta y,z,t)-2u(x,y,z,t)+u(x,y-\Delta y,z,t)}{\Delta y^2} + \frac{u(x,y,z+\Delta z,t)-2u(x,y,z,t)+u(x,y,z-\Delta z,t)}{\Delta z^2} \right) \tag{p}$$

When using the second-order central difference method of calculation, the initial excess pore water pressure of each node in the consolidation area at $t=0$ is known as condition $u_0$. The excess pore water pressure of each node on the boundary can be considered under the following principles:

(a) where the nodes on the drainage boundary surface, the excess pore water pressure at any moment $t$ is 0;
(b) where an impermeable boundary surface is encountered, a point should be set outside the boundary imaginary, so that its $u$ is always equal to the $u$ of the adjacent nodes on the impermeable surface.

These provisions are to meet the boundary conditions on the permeable interface when $t>0, u=0$, and on the impermeable boundary $\frac{\partial u}{\partial n}=0$.

When employing the display method, it is crucial to ensure the fulfilment of the convergence stability condition, which is commonly known as Von-Neumann stability analysis (Charney et al., 1950), and the condition for stability is set as $\frac{C_v \Delta t}{\Delta x^2+\Delta y^2+\Delta z^2} < \xi(n)$ to keep the convergence of three-dimensional PDE, in which $\xi(n)$ denotes the critical convergence factor and $n$ indicates the number of spatial dimensions. This implies that FDMs are unable to accurately capture the physical process when using a large time step size, since the iterations will crash directly as $\frac{C_v \Delta t}{\Delta x^2+\Delta y^2+\Delta z^2} > \xi(n)$. In order to meet the convergence condition and ensure the accuracy of the reference solution, a sufficient number of time intervals $N_t = 21600$ is selected, which means $\Delta t = 4.630 \times 10^{-5}$. This approach is an explicit method that possesses its convergence stability condition. However, if absolute convergence of the numerical method is necessary, an implicit solution method can be adopted instead. Furthermore, in cases where higher precision and efficiency are desired, alternative methods such as the modified Euler method or the Runge-Kutta method can be utilized. The numerical method can be adopted to calculate the exact values $u(x^i, y^i, z^i, t^i)$ of the partial differential equation in the defined domain, which is used to compare between the model-predicted solution and the reference solution when calculating the relative $L_2$ error in three-dimensional cases.

# 5. Physics-informed Deep Learning for Three-dimensional Consolidation

## 5.1 Summary

First, the three-dimensional model is presented to test and validate the proposed PINN framework and as a fundament to solve multidirectional cases. In the three-dimensional model, there are four input neurons representing the spatial and temporal variables $x$, $y$, $z$ and $t$. The output neurons are all one, i.e., the solution value $u$ of the PDE. By comparing the PINN solution and the reference solution, the error of the PINN solution can generally reach $10^{-3}$ orders of magnitude under different working conditions without deliberately tuning the model or improving the algorithm, which is an acceptable error range. In the subsequent sections, details on how to set PINNs for three-dimensional cases in data-driven solution of PDE and data-driven discovery of coefficient of consolidation are presented, different multidirectional drainage conditions are considered in forward problems, data noise influence is analyzed in inverse problems, as well as analysis of PINNs performance, and the problems encountered by PINNs in solving three-dimensional time-dependent PDEs are highlighted.

## 5.2 Forward problems

The governing equation and conditions for three-dimensional consolidation problems are given below:

$$\begin{cases} \dfrac{\partial u}{\partial t} = C_v(\dfrac{\partial^2 u}{\partial x^2} + \dfrac{\partial^2 u}{\partial y^2} + \dfrac{\partial^2 u}{\partial z^2}); \ C_v = 0.05 m^2/a \\ u(x,y,z,0) = p_0 \end{cases} \quad (q)$$

Four typical cases of multidirectional drainage are considered and given in Table 1.

Table 1 Two cases in three-dimensional Terzaghi's consolidation problems.

| No. | $BC$ | Description |
|---|---|---|
| Case 1 | $\begin{cases} u(0,y,z,t) = u(l,y,z,t) = 0 \\ u(x,0,z,t) = u(x,b,z,t) = 0 \\ u(x,y,0,t) = 0; u(x,y,h,t) = 0 \end{cases}$ | Drained boundaries |
| Case 2 | $\begin{cases} u(0,y,z,t) = u(l,y,z,t) = 0 \\ u(x,0,z,t) = u(x,b,z,t) = 0 \\ \dfrac{\partial u(x,y,0,t)}{\partial z} = \dfrac{\partial u(x,y,h,t)}{\partial z} = 0 \end{cases}$ | Undrained top and bottom boundaries |
| Case 3 | $\begin{cases} \dfrac{\partial u(0,y,z,t)}{\partial x} = \dfrac{\partial u(l,y,z,t)}{\partial x} = 0 \\ \dfrac{\partial u(x,0,z,t)}{\partial y} = \dfrac{\partial u(x,b,z,t)}{\partial y} = 0 \\ \dfrac{\partial u(x,y,0,t)}{\partial z} = 0; u(x,y,h,t) = 0 \end{cases}$ | Drained top boundary |

Case 4 $\begin{cases} \dfrac{\partial u(0,y,z,t)}{\partial x} = \dfrac{\partial u(l,y,z,t)}{\partial x} = 0 \\ u(x,0,z,t) = u(x,b,z,t) = 0 \\ \dfrac{\partial u(x,y,0,t)}{\partial z} = \dfrac{\partial u(x,y,h,t)}{\partial z} = 0 \end{cases}$  drained front and back boundaries where $p_0 = 1\ Pa$ is the initial excess pore water pressure, $h = 1\ m$ is the thickness of the soil layer, $l = 1\ m$ is the length of the soil layer, and $b = 1\ m$ is the width of the soil layer. Therefore, $0 < t \leq 1\ year$, $0 < x \leq l$, $0 < y \leq b$, and $0 < z \leq h$ are considered for the initial and boundary conditions. The three-dimensional model builds upon the one-dimensional model by incorporating additional spatial dimensions, resulting in a neural network with four input neurons, representing three spatial dimensions and one temporal dimension. This model has six different boundary sides that indicate the multidirectional drainage of three-dimensional soil consolidation, the drained boundary corresponds to the Dirichlet boundary condition while the undrained boundary is to the Neumann boundary condition.

In this section, the physics-informed deep learning model is set and trained to solve 3D Terzaghi's consolidation PDE by using the collocation points randomly selected from the whole domain based on the initial and boundary conditions, the loss items of the loss function differ slightly from those in the previous section due to the disparity in boundary conditions along the $x$ and $y$ directions compared to the $z$ direction, the inclusion of more loss items creates additional difficulties in the balance between different items of the loss function, which in turn results in higher computational time and more difficult convergence. For the PINNs applied to three-dimensional cases, the loss function is minimized by the L-BFGS algorithm through the training process of the model, which shows a significant decline and approaches zero, the learning rate of the L-BFGS optimizer is set as 1.0 to ensure better convergence, The initial condition, boundary conditions and physical constraints are coded into the loss function to make the outputs better consistent with real physics when the loss function converges to 0, and the MSE loss function in this section is initially defined below according to section 3.2:

$$MSE = \omega_{ui} MSE_{ui} + \omega_{ux} MSE_{ux} + \omega_{uy} MSE_{uy} + \omega_{uz} MSE_{uz} + \omega_f MSE_f \qquad (r)$$

In Equation (r), $MSE_{ui}$ is the mean squared error of initial condition, $MSE_{ux}$, $MSE_{uy}$ and $MSE_{uz}$ are the mean squared error from $X$ boundary, $Y$ boundary and $Z$ boundary respectively, and $MSE_f$ represents the PDE loss. The items in front of each $MSE$ item are their corresponding loss weight parameters, which are all initially set as 1. We generate 19 time steps and 35 sample points in both the $x$, $y$ and $z$ coordinates, so there are 814625 solution sample points in the whole domain. The number of training points from initial and boundary conditions is set as $N_u = 80000$ while the number of collocation points is defined as $N_f = 482525$, including 300000 inner domain points and all the 182525 initial and boundary points. Furthermore, the model incorporates 4 hidden layers with 40 neurons each. Due to the large amount of data, if the data cannot pass through the neural network in one go, we strongly recommend setting batches to increase the training speed and optimization of the model. Figure 3 presents the diagrams illustrating the trend of the Mean Squared Error (MSE) loss function after the model processing.

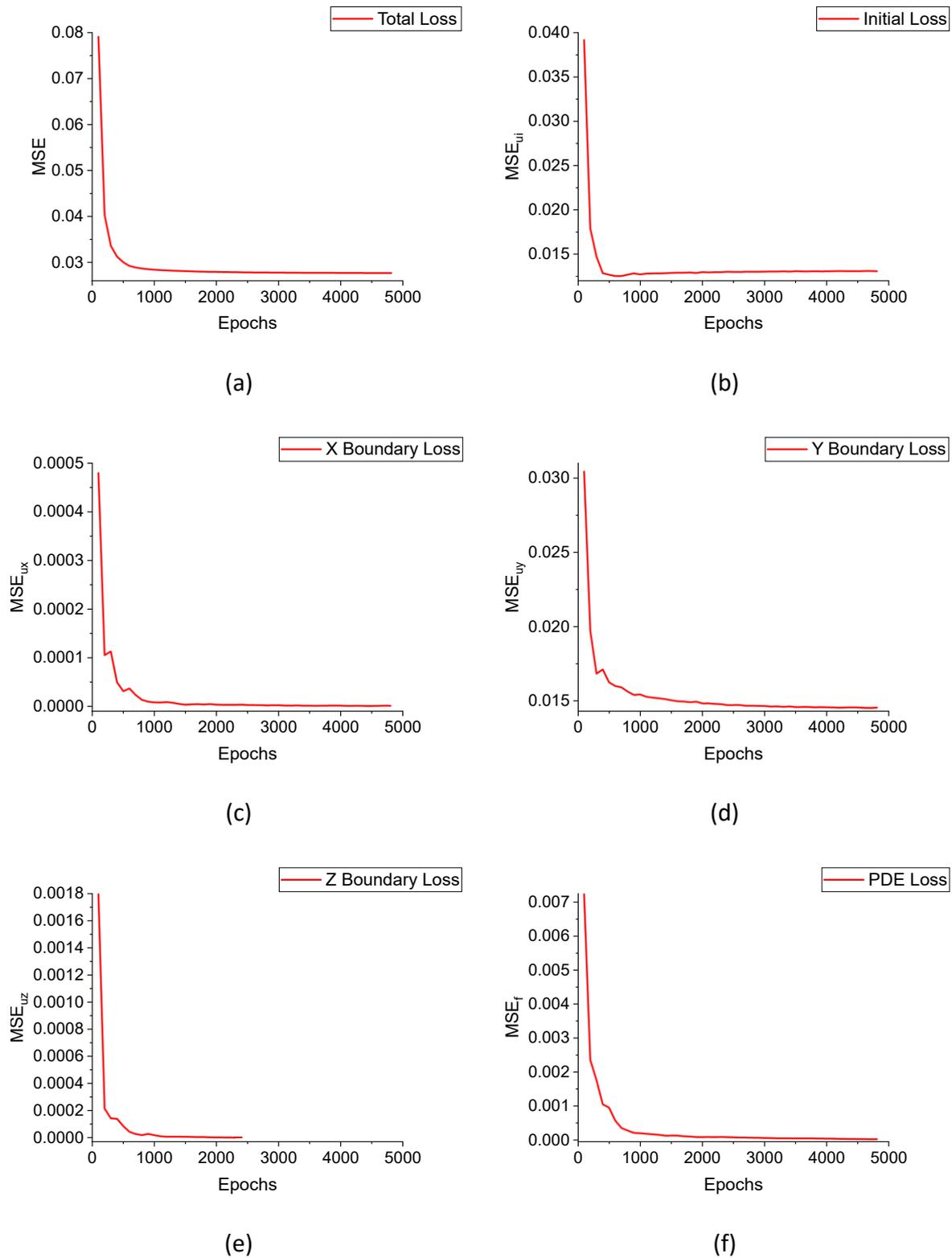

(a)

(b)

(c)

(d)

(e)

(f)

**Figure 3** The trend of the Mean Squared Error (MSE) loss function for three-dimensional Terzaghi's consolidation problem case 4 in the physics-informed deep learning model, in terms of (a) Total loss, (b) Initial loss, (c) boundary loss in the x-direction, (d) boundary loss in the y-direction, (e) boundary loss in the z-direction and (f) PDE loss.

To make the PINNs work better in three-dimensional settings, a normal initialization is used before model training. The total training time for case 4 is 1625s after 5000 epochs, and as shown in Figure 3, the loss of the initial condition, each boundary and PDE of the three-dimensional model present a fine downward trend and gradually stabilize after 2000 epochs of training. The final $MSE$ of $x$, $z$ boundary and PDE decline to $4.024 \times 10^{-6}$, $3.813 \times 10^{-6}$ and $2.901 \times 10^{-5}$, respectively, which means the difference between the predicted and true values is at a desirable small level. This indicates that physics-informed deep learning has a good performance for solving three-dimensional consolidation PDE.

### 5.2.1 Drainage on Four to Six Sides

In this section, multidirectional drainage on four to six sides is considered. To facilitate quantitative comparisons between the PINN solution and the reference solution, three specific time points have been carefully selected ($t = 0.20$, $0.50$ and $1.0$). The results obtained for each solution are displayed in Figure 4, allowing for a comprehensive evaluation and analysis of their respective performances. The color plots display the prediction results from the model for case 1 and case 2 by utilizing an interpolation method based on the excess pore water pressure values of the nearest points. The heat maps shown in Figure 4 are only for visualization purposes, which shows the effect of the deep learning output.

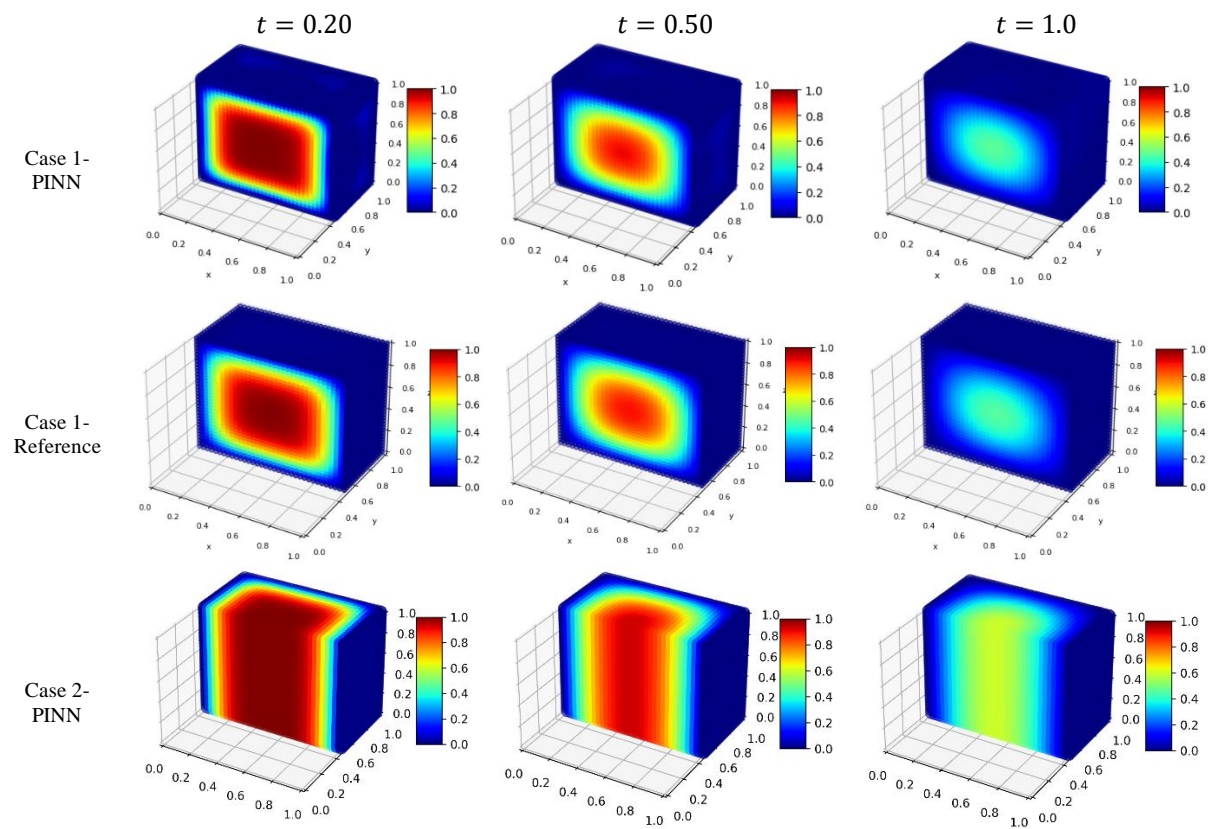

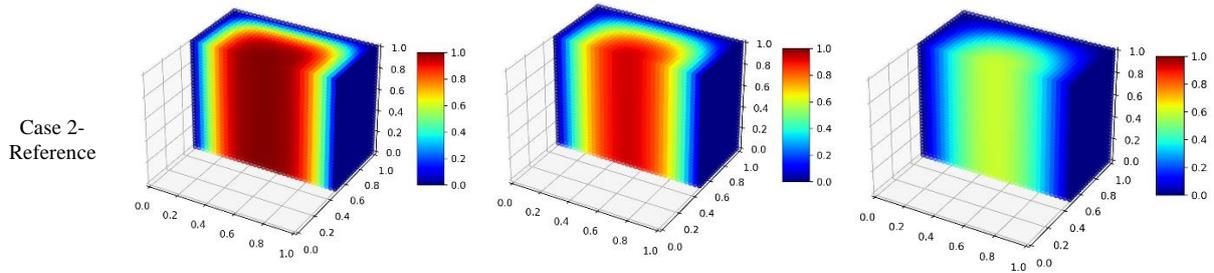

Case 2-Reference

**Figure 4** The heat maps of the results of three-dimensional Terzaghi's consolidation problems predicted by the physics-informed deep learning model and numerical model considering two cases (case 1 and case 2).

The results indicate that the excess pore water pressure in the soil layer changes with time, and the consistency between the PINN solution and the reference solution is excellent at different time points. For better comparisons, a numerical solution to this problem is shown in Figure 5 as a reference to show the PINN performance. Three time points ($t_0 = 0.20, t_1 = 0.40, t_2 = 0.80$) are randomly selected to make comparisons quantitatively between the PINN prediction and reference solution at these time points, the results are shown in Figure 5.

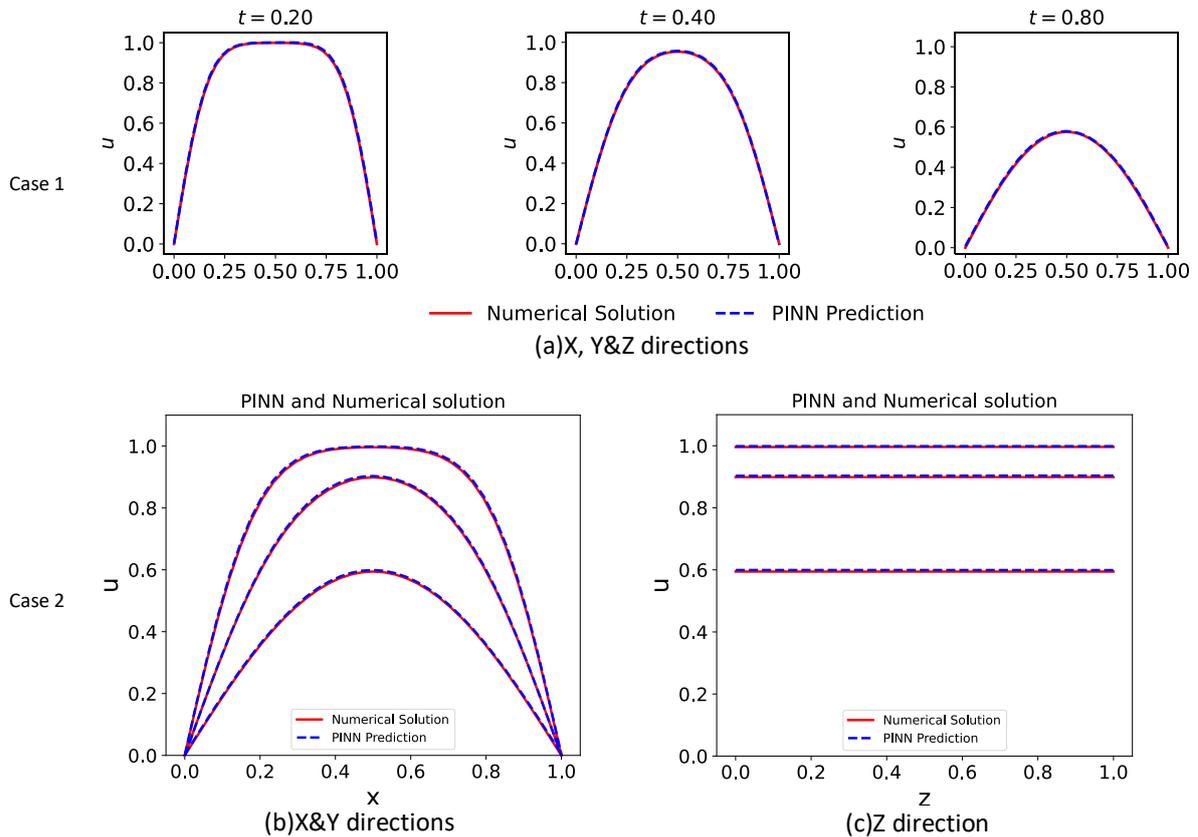

Case 1

Case 2

(a) X, Y&Z directions

(b) X&Y directions

(c) Z direction

**Figure 5** The comparisons of three-dimensional Terzaghi's consolidation problems between the physics-informed deep learning model and numerical model at three selected time points $t_0 = 0.20, t_1 = 0.40, t_2 = 0.80$. (a) case 1-X, Y&Z directions, (b) case 2-X&Y directions and (c) case 2-Z direction. The excess pore water pressure decreases over time in the whole domain.

To quantify the difference between the predicted solution and reference solution, the $L_2$ norm of the relative error and the Mean Absolute Error (MAE) are introduced and expressed as follows:

$$e = \frac{||\hat{u}-u||_{L_2}}{||u||_{L_2}} \tag{i}$$

$$MAE = \frac{|\hat{u}-u|}{u} \tag{ii}$$

Figure 5 intuitively indicates that the PINN solution and the reference solution match well. The MAE in the whole domain is $9.065 \times 10^{-3}$ and $9.015 \times 10^{-3}$ in case 1 and case 2, respectively, which indicates that the PINN solution agrees well with the reference solution. The results summarised in Table 2 demonstrate that PINNs can achieve comparable accuracy as the numerical solution in the magnitude of the relative $L_2$ error. Moreover, the prediction time of PINNs in this section is 0.52s, which is significantly less than the inference time of the reference solution which is 54s. This shows that the PINNs method is fast and equally accurate.

**Table 2** The Relative $L_2$ Errors between PINN prediction and numerical solution for three-dimensional Terzaghi's consolidation problems.

| time | $t = 0.20$ | $t = 0.40$ | $t = 0.80$ |
|---|---|---|---|
| Relative $L_2$ Error-Case 1 | $9.264 \times 10^{-3}$ | $9.711 \times 10^{-3}$ | $9.569 \times 10^{-3}$ |
| Relative $L_2$ Error-Case 2 | $9.581 \times 10^{-3}$ | $7.781 \times 10^{-3}$ | $8.645 \times 10^{-3}$ |

To analyze our method's performance in 3D consolidation problems, we conduct the sensitivity analysis to measure its prediction accuracy for various neural network architectures. Taking Case 2 as an example, Table 3 displays the errors for varying numbers of hidden layers and neurons per layer. As anticipated, the accuracy of predictions increases with the number of layers and neurons, allowing the neural network to approximate more complex functions.

**Table 3** The Errors between PINN prediction and numerical solution for different number of hidden layers and different number of neurons per layer for three-dimensional Terzaghi's consolidation problems (Case 2).

| Layers | Neurons | 10 | | | 20 | | | 40 | | |
|---|---|---|---|---|---|---|---|---|---|---|
| | time | t=0.20 | t=0.40 | t=0.80 | t=0.20 | t=0.40 | t=0.80 | t=0.20 | t=0.40 | t=0.80 |
| 1 | $L_2$ | 2.1e-01 | 2.0e-01 | 1.9e-01 | 1.9e-01 | 1.4e-01 | 8.8e-02 | 1.7e-01 | 1.0e-01 | 6.5e-02 |
| | MAE | 9.3e-02 | | | 6.6e-02 | | | 5.5e-02 | | |
| 2 | $L_2$ | 9.7e-02 | 4.8e-02 | 5.1e-02 | 4.4e-02 | 2.3e-02 | 2.6e-02 | 2.9e-02 | 2.1e-02 | 1.7e-02 |
| | MAE | 3.6e-02 | | | 2.2e-02 | | | 1.8e-02 | | |
| 4 | $L_2$ | 2.7e-02 | 1.4e-02 | 1.6e-02 | 1.1e-02 | 8.2e-03 | 7.6e-03 | 9.6e-03 | 7.8e-03 | 8.7e-03 |
| | MAE | 1.6e-02 | | | 1.1e-02 | | | 9.0e-03 | | |

### 5.2.2 Drainage on One to Two Sides

In this section, we consider the drainage on one to two sides for the three-dimensional consolidation model. To facilitate quantitative comparisons between the PINN solution and the reference solution, three specific time points have been carefully selected ($t = 0.20$, $0.50$ and 1.0). The results obtained for each solution are displayed in Figure 6, allowing for a comprehensive evaluation and analysis of their respective performances.

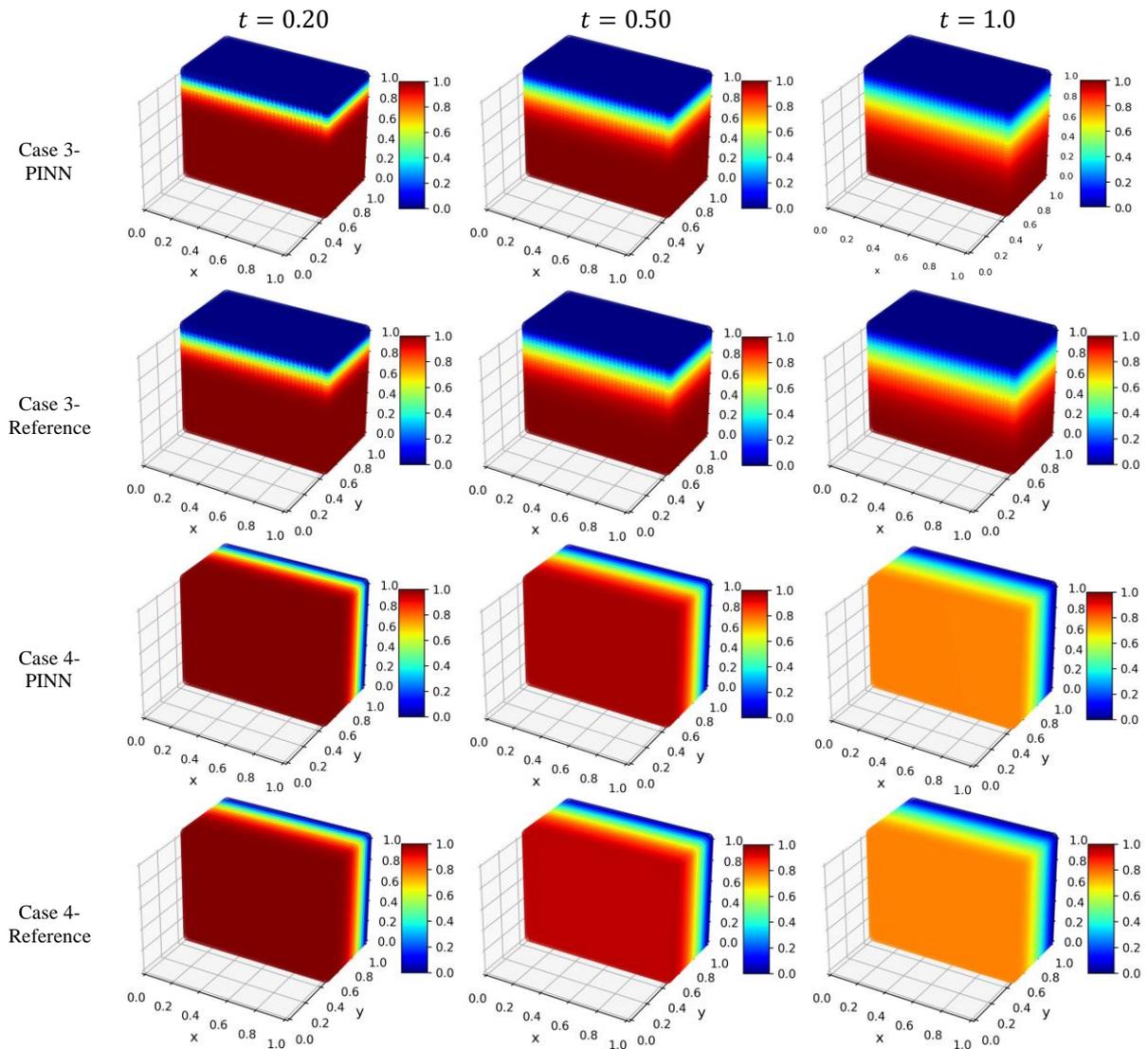

**Figure 6** The heat maps of the results of three-dimensional Terzaghi's consolidation problems predicted by the physics-informed deep learning model and numerical model considering two cases (case 3 and case 4).

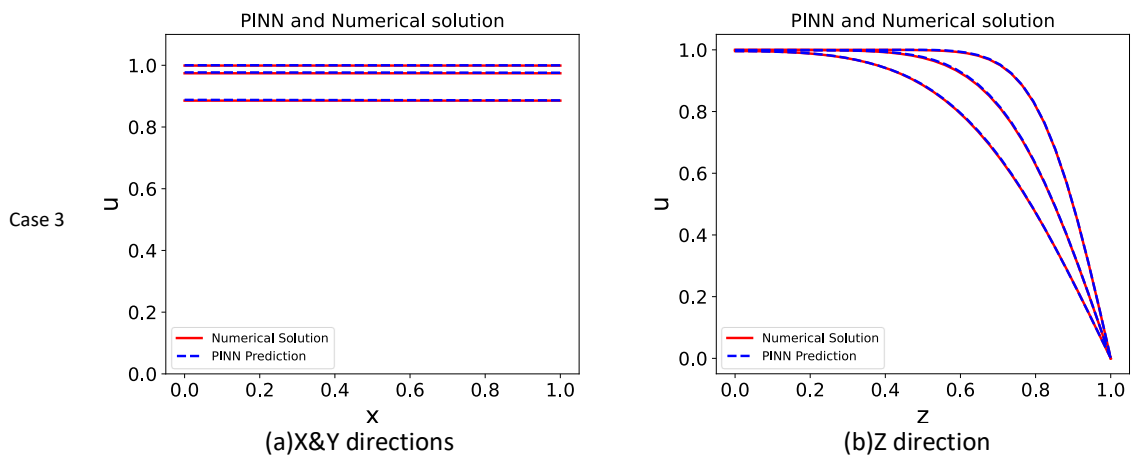

(a) X&Y directions  (b) Z direction

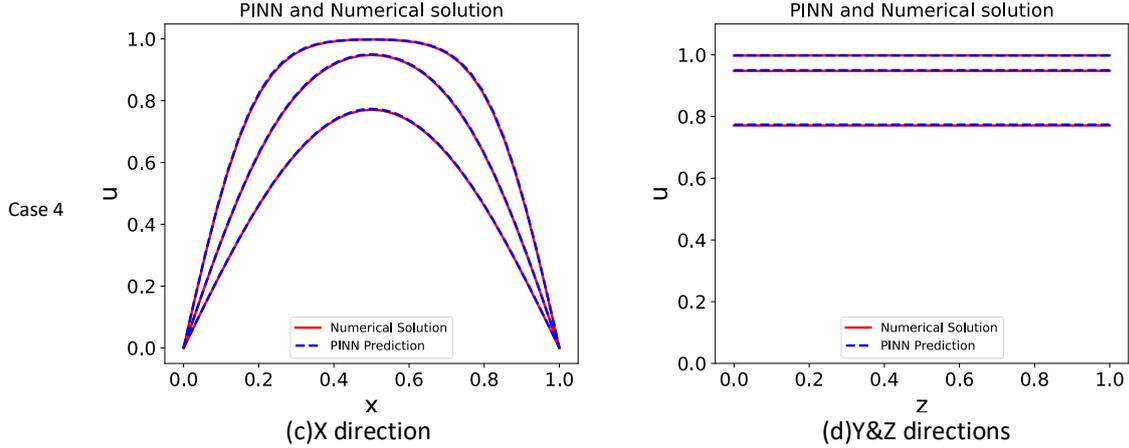

(c)X direction                (d)Y&Z directions

**Figure 7** The comparisons of three-dimensional Terzaghi's consolidation problems between the physics-informed deep learning model and numerical model at three selected time points $t_0 = 0.20, t_1 = 0.40, t_2 = 1.0$. (a) case 3-X&Y directions, (b) case 3-Z direction, (c) case 4-X direction and (d) case 4-Y&Z directions. The excess pore water pressure decreases over time in the whole domain.

Figure *6* and Figure *7* indicate that the PINN solution is consistent well with the numerical solution, and the comparisons are quantified by relative $L_2$ error, which is shown in Table 4. The Mean Absolute Error (MAE) in the whole domain is $8.669 \times 10^{-3}$ and $4.456 \times 10^{-3}$ in case 3 and case 4, respectively, which indicates that the PINN solution agrees well with the reference solution. Moreover, the prediction time of PINNs in this section is 0.46s while the inference time of reference solution is 52s. This indicates that there are substantial advantages in the use of PINN solutions for three-dimensional consolidation problems.

**Table 4** The Relative $L_2$ Errors between PINN prediction and numerical solution for three-dimensional Terzaghi's consolidation problems.

| time | $t = 0.20$ | $t = 0.40$ | $t = 0.80$ |
|---|---|---|---|
| Relative $L_2$ Error-Case 3 | $1.325 \times 10^{-3}$ | $1.284 \times 10^{-3}$ | $1.057 \times 10^{-3}$ |
| Relative $L_2$ Error-Case 4 | $4.783 \times 10^{-3}$ | $3.170 \times 10^{-3}$ | $2.114 \times 10^{-3}$ |

The PINN solution agrees well with the numerical solution. In traditional numerical methods (FEMs and FDMs), the domain of the problem needs to be discretized in a mesh whereas the PINNs method is meshfree. Undoubtedly, the quality of the mesh will influence the convergence and accuracy of FEMs and FDMs. In fact, if the spatial step is too small or the time step is too large, the convergence of the numerical model will be poor, which can also be verified by evidence in section 4.1.

To further analyze the performance of our method in 3D consolidation problems, we perform the following sensitivity analysis to quantify its prediction accuracy for different neural network architectures. Taking Case 3 as an example, Table 5 shows the errors for different numbers of hidden layers and different numbers of neurons per layer. As expected, we observe that the prediction accuracy improves as the number of layers and the number of neurons increases, and thus the neural network can approximate more complex functions.

**Table 5** The Errors between PINN prediction and numerical solution for different number of hidden layers and different number of neurons per layer for three-dimensional Terzaghi's consolidation problems (Case 3).

| Layers | Neurons | 10 | | | 20 | | | 40 | | |
|---|---|---|---|---|---|---|---|---|---|---|
| | time | t=0.20 | t=0.40 | t=0.80 | t=0.20 | t=0.40 | t=0.80 | t=0.20 | t=0.40 | t=0.80 |
| 1 | $L_2$ | 3.3e-02 | 2.2e-02 | 1.0e-02 | 2.1e-02 | 8.0e-03 | 7.9e-03 | 2.3e-02 | 9.5e-03 | 6.0e-03 |
| 1 | MAE | 1.9e-02 | | | 1.6e-02 | | | 1.6e-02 | | |
| 2 | $L_2$ | 1.6e-03 | 1.9e-03 | 2.9e-03 | 3.2e-03 | 1.7e-03 | 3.6e-03 | 2.6e-03 | 2.8e-03 | 4.6e-03 |
| 2 | MAE | 9.8e-03 | | | 1.0e-02 | | | 1.0e-02 | | |
| 4 | $L_2$ | 1.4e-03 | 2.0e-03 | 1.8e-03 | 9.3e-04 | 1.1e-03 | 2.6e-03 | 1.3e-03 | 1.3e-03 | 1.1e-03 |
| 4 | MAE | 8.8e-03 | | | 8.9e-03 | | | 8.7e-03 | | |

### 5.3 Inverse Problems
#### 5.3.1 Identification of Coefficient of Consolidation

In the previous sections, it was shown that the PINNs method can solve three-dimensional consolidation forward problems with good performance. The accuracy achieved for the PINNs indicates that the excess pore water pressure in the soil layer can be determined directly by physics-informed deep learning provided that the initial and boundary conditions are known. It does not require having a deep prior knowledge of specific physics in soil mechanics. While FDMs and FEMs are also good for forward problems, these methods are not very convenient to address inverse problems using classical numerical methods.

In this section, the PINNs method is used to solve inverse problems for case 1 in three-dimensional consolidation (Equation (a)). The solution of the 3D numerical model is used as the exact solution, and the coefficient of consolidation $C_v$ is unknown. Thus, we can train the PINNs to obtain the unknown coefficient by incorporating both the exact solution and governing physics into the neural network. The MSE loss function for inverse problems is considered as follows based on Equations (h)(i)(k):

$$MSE = \frac{1}{N_u}\sum_{i=1}^{N_u}|\hat{u}_i - u_i|^2 + \frac{1}{N_u}\sum_{i=1}^{N_u}\left|\frac{\partial \hat{u}_i}{\partial t} - C_v(\sum \frac{\partial^2 \hat{u}_i}{\partial n^2})\right|^2 \qquad \text{(iii)}$$

The number of training data ($N_u$) randomly selected from the exact solution is used to compute the MSE, $\hat{u}_i$ is the approximated solution of the PDEs predicted by PINNs, and $u_i$ is the exact values of the PDEs solution, $n$ indicates the spatial dimensions and coordinates $x, y, z$. The unknown $C_v$ is included in the expression for MSE loss that needs to be discovered by the PINNs.

The exact coefficients of consolidation $C_v$ in three-dimensional $C_v$ identification problems are assumed as 0.01, 0.05 and 0.1 respectively with all drained boundaries (inverse problems for case 1). The depth of PINNs is 5 and the number of hidden layers is 4 with 40 neurons in each layer, and the activation function employed at the hidden layers is Tanh$(x)$. The L-BFGS optimizer with a learning rate of 1.0 and the Adam optimizer with a learning rate of 0.001 are used for inverse problems in three-dimensional consolidation, we use Adam algorithm for 10000 epochs first and then use L-BFGS algorithm to ensure the computational efficiency and accuracy. The batch size of the training data points is 4000 with 1% noise. The parameter trained to approximate the exact coefficient is initialized to 1.0 in the proposed PINN framework.

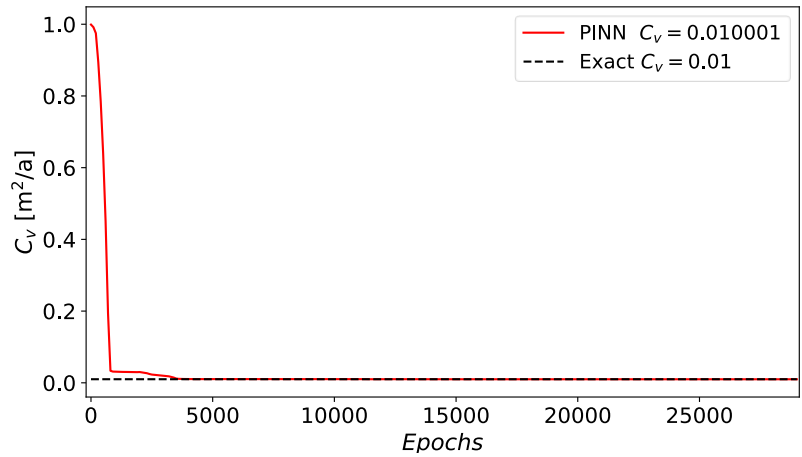

(a) 3D $C_v$ identification problem ($C_v = 0.01$).

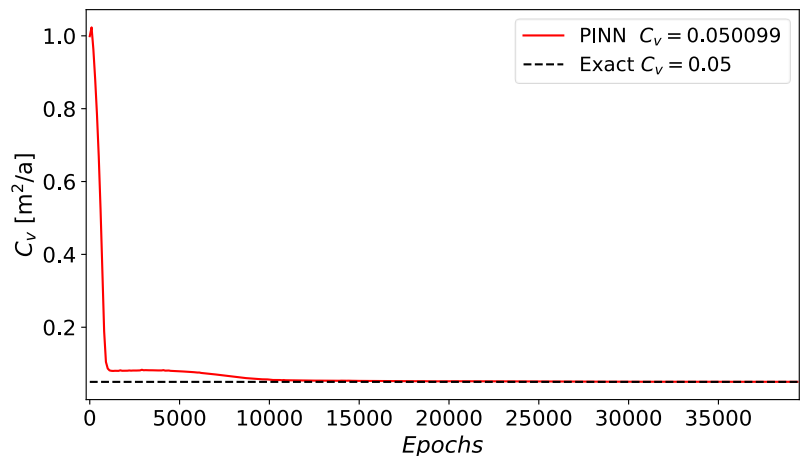

(b) 3D $C_v$ identification problem ($C_v = 0.05$).

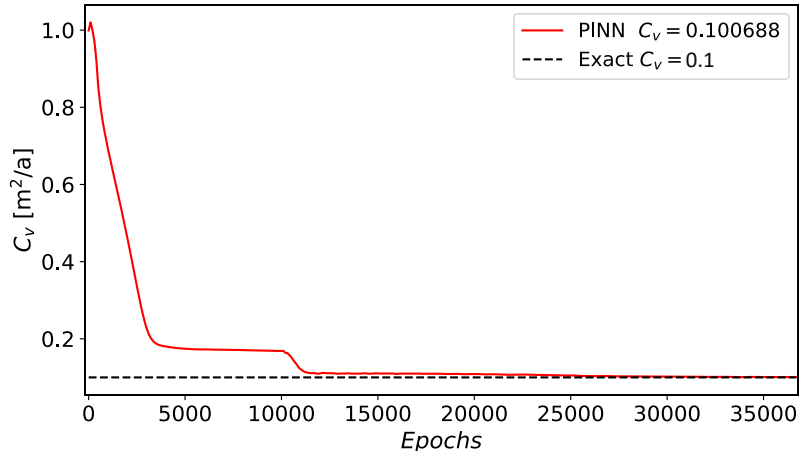

(c) 3D $C_v$ identification problem ($C_v = 0.1$).

**Figure 8** The evolution of the PINN-predicted coefficient of consolidation in three-dimensional $C_v$ identification problems (a) $C_v = 0.01$, (b) $C_v = 0.05$ and (c) $C_v = 0.1$.

The accuracy of PINNs for three-dimensional inverse problems can be identified by the relative error $E$ and accuracy $A$, which are defined as:

$$E = \frac{|C_v^p - C_v^e|}{C_v^e} \tag{iv}$$

$$A = (1 - E) \times 100\% \tag{v}$$

where $C_v^e$ and $C_v^p$ are the exact and PINN-predicted coefficient of consolidation $C_v$. Figure 8 shows the results of PINN-predicted coefficient of consolidation $C_v$ in three-dimensional inverse problems of soil consolidation. PINNs can predict excellent results in $C_v$ parameter identification in three-dimensional consolidation inverse problems (Table 6) with relative errors of $7.00 \times 10^{-5}$, $1.98 \times 10^{-3}$ and $6.88 \times 10^{-3}$, respectively. It is worth mentioning that as the value of $C_v$ increases, the accuracy of the prediction of the coefficient $C_v$ slightly decreases. Moreover, the number of epochs required for model training to achieve an acceptable error increases, with 4,000, 10,000, and 25,000 for the $C_v = 0.01$, $C_v = 0.05$, and $C_v = 0.1$ inverse problems, respectively. This indicates that it is more difficult to achieve convergence when solving three-dimensional and multidirectional problems by PINNs and requires more computing time and computing power. However, the proposed PINN method shows very good performance in inverse problems for three-dimensional consolidation.

**Table 6** The PINN-predicted coefficient of consolidation $C_v$ and corresponding Relative Errors (REs) and accuracies for three-dimensional Terzaghi's consolidation problems.

| dimension | 3D | 3D | 3D |
|---|---|---|---|
| Exact $C_v$ | 0.01 | 0.05 | 0.10 |
| Predicted $C_v$ | 0.0100007 | 0.050099 | 0.100688 |
| Error | $7 \times 10^{-5}$ | $1.98 \times 10^{-3}$ | $6.88 \times 10^{-3}$ |
| Accuracy (%) | 99.993 | 99.802 | 99.312 |

Table 6 presents that the average accuracy of $C_v$ coefficient inversion in three-dimensional settings is over 99.7%. In conclusion, the proposed PINN framework is verified to be effective and efficient in data-driven discovery of PDE for three-dimensional and multidirectional problems.

### 5.3.2 Noise Influence Analysis

In addition, the data input to the neural network for parameter inversion is not always accurate, and the data is often noisy, i.e. observed data or experimental data. Zhang et al. (2022) and Zhang et al. (2023) carried out the noise influence analysis of inverse problems in one-dimensional soil consolidation, but it is not clear what the influence of noisy data in the identification of coefficient of consolidation for three-dimensional and multidirectional consolidation problems. Based on the PINNs settings in the last section, an investigation of noise influence analysis to evaluate the performance of PINNs under interference in three-dimensional inverse problems is conducted, and the noisy data is created from the exact data according to Gaussian distribution:

$$u_{noise} = u_{exact} + \varepsilon \cdot STD(u_{exact}) \cdot N(0,1) \tag{vi}$$

where $u_{exact}$ and $u_{noise}$ are the vector of the exact data and the noisy data, respectively; $\varepsilon$ is the Gaussian noise level (%); $STD(\cdot)$ is a function used to figure out the standard deviation of the exact data, and $N(0,1)$ is a randomly generated vector conforming to the Gaussian distribution.

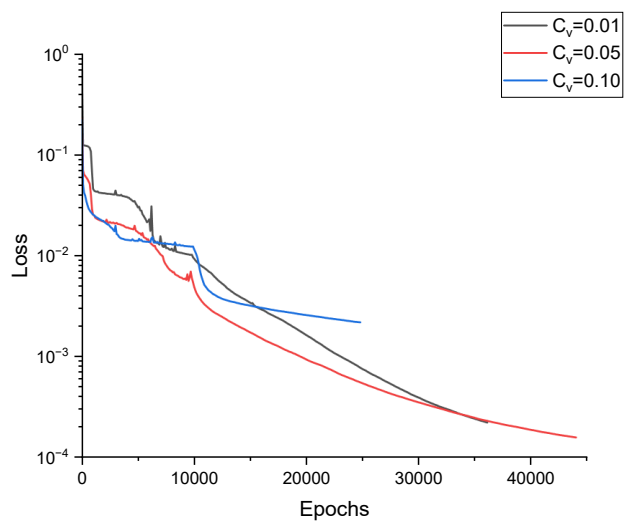

**Figure 9** The evolution of loss value in three-dimensional $C_v$ identification problems with noise-free data

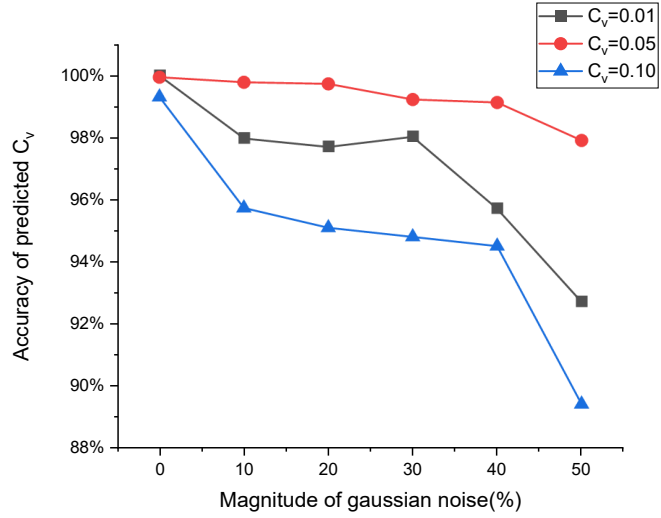

**Figure 10** The influence of gaussian noise level on the accuracy in three-dimensional $C_v$ identification problems

As is demonstrated in Figure 9 and Figure 10, the more the loss function decreases, the higher the accuracy of $C_v$ identification. It is also obvious to note that the accuracies of the identified coefficients decline as the Gaussian noise level increases: when the noise level is less than 30%, the accuracy is above 95%, whereas for the 50% noise level, there is the lowest accuracy. Nevertheless, the proposed PINN method can still accurately identify the unknown parameters from the noisy data within an acceptable accuracy for three-dimensional cases.

### 5.4 Soil Settlement Prediction

The consolidation process represented in Equation (a) describes the settlement and consolidation of the soil layer. The degree of consolidation is a parameter employed to quantify the extent of settlement of the soil layer at a given time, typically expressed as a percentage denoted by $U(\%)$:

$$U = \frac{S_t}{S} \tag{s}$$

where $S_t$ represents the settlement at time $t$ and $S$ is the final settlement of the soil layer, which are denoted by Equation (t) and (u), respectively. The vertical settlement ($z$-direction) in the center of the top side of the soil layer is considered in this section. It is assumed that the thickness of the soil layer is $h$, and the load is a one-time applied full load $p_0$.

$$S_t = m_v(p_0 - u_m)h \tag{t}$$

$$S = m_v p_0 h \tag{u}$$

$u_m$ is the excess pore water pressure through the thickness $h$, and $m_v$ is the coefficient of volume compression of the soil layer, and they are defined in Equations (v) and (w), respectively.

$$u_m = \frac{1}{h}\int_0^h u\,dz \qquad (v)$$

$$m_v = \frac{a_v}{1+e_0} \qquad (w)$$

Case 1 is taken as an example in this problem, which means all the sides are drained. The coefficient of volume and initial porosity are set as $a_v = 0.00025\,kpa^{-1}$ and $e_0 = 0.8$, respectively. Based on Equations (t)(v)(w), the $S(t) - t$ relationship can be represented by combining the mathematical theory and the predictions obtained from the PINNs. The results are depicted in Figure 11, where the red solid line represents the reference solution, and the blue dashed line corresponds to the PINN solution. In addition, the degree of consolidation $U(\%)$ of 70% after a year can be computed using Equation (s). The findings substantiate a strong agreement between the soil settlement predicted by the PINNs and the reference analysis with a MAE of $7.2 \times 10^{-3}$ (accuracy: 99.28%). This underscores the efficacy of the PINNs method in accurately capturing the complex process of soil consolidation. Therefore, quick prediction of soil settlement can be made repeatedly by the proposed PINNs method after being trained well.

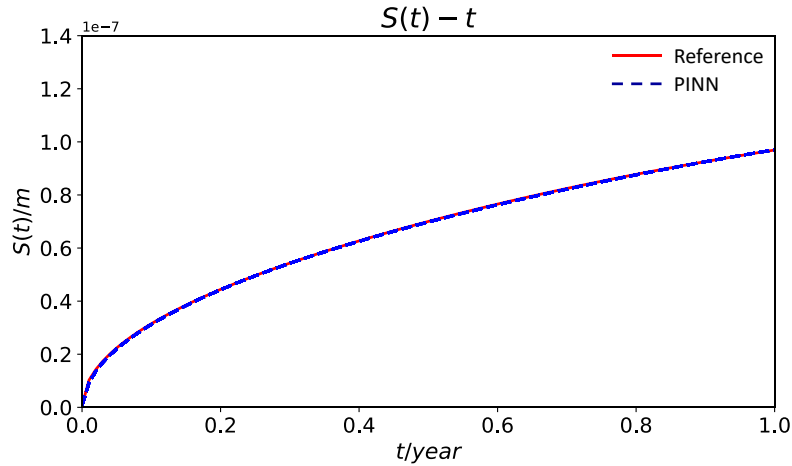

**Figure 11** The graph of $S(t) - t$ relationship comparison between PINN solution (blue dashed line) and reference solution (red solid line).

## 6. Conclusions

In this paper, a physics-informed neural networks (PINNs) framework for predicting three-dimensional soil consolidation problems is proposed. The predictions capture the excess pore water pressure, the coefficient of consolidation and soil settlement. Several initial and boundary conditions are considered to evaluate the performance of the proposed PINN method under multidirectional cases. The three-dimensional Terzaghi's consolidation forward and inverse problems are then solved by using both the proposed PINN method and numerical methods and their performance is compared. The results show that:

(1) The PINNs can accurately solve three-dimensional partial differential equations (with four independent variables) employing a simpler and more direct approach compared with the traditional numerical methods that rely on complex theoretical derivations;

(2) Unlike traditional numerical analysis methods such as Finite Element Methods (FEMs) and Finite Difference Methods (FDMs), which are based on numerical discretization to tackle and approximate the three-dimensional PDEs, the PINNs method is mesh-free. Thus, its solution process is not constrained by the mesh division, nor the mesh division influences its convergence, which is one of the main shortcomings of existing FEMs and FDMs. In addition, we also do not need a lot of prior mathematical and physical knowledge;

(3) In three-dimensional forward problems, PINN training can be achieved with a smaller set of training points in the whole domain, and fast prediction can be reached with a relatively short computation time compared with numerical analysis;

(4) The loss function decreases through the training process of the deep learning network to obtain a solution that meets the error tolerance. In fact, the PINN solution is in good agreement with the reference solution with the error of order $10^{-3}$;

(5) As a deep learning method, embedding sparse data in PINNs can accelerate its convergence and improve its accuracy. In three-dimensional inverse problems, the proposed PINNs method can be used to identify the coefficient of consolidation $C_v$, and the MAEs are in the order lower than $10^{-3}$ with 1% noisy training data. The proposed PINN framework shows good performance in parameter identification problems with an accuracy of over 99.3%.

(6) The noisy data has a significant impact on the accuracy of parameter identification, but the proposed PINN method can still achieve a desirable accuracy in three-dimensional cases.

However, there are still improvements to be made in PINNs for solving large and complex, three-dimensional or multidirectional problems, there are:

(1) The computational time required for PINNs to solve three-dimensional equations is higher than that for the one-dimensional cases, so the computing resource is a requirement for algorithms optimization;

(2) The accuracy of PINNs solution for three-dimensional problems is more challenging and typically the error is generally in the order of $10^{-3}$;

(3) For inverse problems, a balance between the data-driven and physical constraints is critical. Theoretically, PINNs can incorporate more experimental or observational data to solve inverse problems that utilize both data-driven and physical constraints.

Finally, while the PINN method proposed in this study is a great step forward in the prediction of consolidation settlement problems from a practical standpoint, further research in PINNs for high-dimensional partial differential equations is needed at both the theoretical and applied levels as there is a lack of sufficient empirical evidence. Nevertheless, the impact of PINNs on engineering research in recent years has been remarkable, bringing new ideas to engineers for solving complex physical problems.

## CRediT authorship contribution statement

## Declaration of competing interest

The authors declare that he has no known competing financial interests or personal relationships that could have appeared to influence the work reported in this paper.

## Acknowledgements

The first author would like to acknowledge the financial support from the China Scholarship Council-University of Leeds Scholarships for his study at the University of Leeds.